\documentclass[twocolumn]{ceurart}
\usepackage{xspace}
\usepackage{longtable}
\usepackage{multirow}
\usepackage{tabularx}
\usepackage{array}
\usepackage{adjustbox}
\usepackage{multicol, blindtext}
\usepackage{listings}
\begin{document}

\copyrightyear{2022}
\copyrightclause{Copyright for this paper by its authors. Use permitted under Creative Commons License Attribution 4.0 International (CC BY 4.0).}

\conference{4th Edition of Knowledge-aware and Conversational Recommender Systems (KaRS) Workshop @ RecSys 2022, September 18--23 2023, Seattle, WA, USA.}
\newcommand{\infact}{\textsc{INFACT}\xspace}
\title{INSPIRED2: An Improved Dataset for Sociable Conversational Recommendation}

\author{Ahtsham Manzoor}[%
orcid=0000-0001-9418-753,
email=ahtsham.manzoor@aau.at,
url=https://ahtsham58.github.io/,
]
\cormark[1]
\address{University of Klagenfurt, Universitätsstraße 65-67, Klagenfurt am Wörthersee, 9020, Austria }

\author{Dietmar Jannach}[%
orcid=0000-0002-4698-8507,
email=dietmar.jannach@aau.at,
url=https://www.aau.at/en/aics/research-groups/infsys/team/dietmar-jannach/,
]

\cortext[1]{Corresponding author.}
\renewcommand{\shortauthors}{A. Manzoor}
\newcommand{\crbCRS}{\textsc{CRB-CRS}\xspace}
\newcommand{\rbCRS}{\textsc{RB-CRS}\xspace}
\newcommand{\inspcurrent}{\textsc{INSPIRED2}\xspace}
\newcommand{\inspmain}{\textsc{INSPIRED}\xspace}
\begin{abstract}
Conversational recommender systems (CRS) that are able to interact with users in natural language often utilize recommendation dialogs which were previously collected with the help of paired humans, where one plays the role of a \emph{seeker} and the other as a \emph{recommender}. These recommendation dialogs include items and entities that indicate the users' preferences. In order to precisely model the seekers' preferences and respond consistently, CRS typically rely on item and entity annotations. 
A recent example of such a dataset is \inspmain, wich consists of recommendation dialogs for \emph{sociable} conversational recommendation, where items and entities were annotated using \emph{automatic} keyword or pattern matching techniques. An analysis of this dataset unfortunately revealed that there is a substantial number of cases where items and entities were either wrongly annotated or annotations were missing at all. This leads to the question to what extent automatic techniques for annotations are effective. Moreover, it is important to study impact of annotation quality on the overall effectiveness of a CRS in terms of the quality of the system's responses.  To study these aspects, we manually fixed the annotations in \inspmain. 
We then evaluated the performance of several benchmark CRS using both versions of the dataset. Our analyses suggest that the improved version of the dataset, i.e., \inspcurrent, helped increase the performance of several benchmark CRS, emphasizing the importance of data quality both for end-to-end learning and retrieval-based approaches to conversational recommendation. 
We release our improved dataset (\inspcurrent) publicly at  \textcolor{blue}{\url{https://github.com/ahtsham58/INSPIRED2}}.
\end{abstract}


\maketitle

\section{Introduction}
Sociable conversational recommender systems (CRS) aim to build rapport with users while interacting with them in natural language \cite{hayati2020inspired,Pecune2020}. 
CRS that rely on natural language processing (NLP) nowadays commonly utilize datasets of previously recorded dialogs between humans, where one plays the role of a \emph{recommendation-seeker} and the other as \emph{human-recommender}, see e.g., \cite{li2018towards}. However, due to a certain lack of rich sociable interactions in such datasets \cite{jannach2022conversational}, it can be challenging to build a sociable CRS that builds rapport with the users using such limited data.

Therefore, it is important to develop datasets like \inspmain \cite{hayati2020inspired}, which includes dialogs that implement rich social communication strategies. Such rich datasets represent a solid basis to develop trustable CRS that are able to engage users in a natural and user-adaptive manner. Another key factor for building high-quality CRS lies in the proper recognition of the named entities and other concepts that appear in the dialogs. In the movies domain, for example, being able to exactly identifying the items (i.e., movies) and related entities and concepts (e.g., actors or genres) can play a pivotal role for building an effective system. Existing CRS for example arrange such entities and their relationships as graphs \cite{di2021dialogue,9458937}, and these relationships often form the basis to model the users' preferences, e.g., \cite{cao2019unifying,chen-etal-2019-towards,zhou2021crfr}.  Moreover, domain specific concepts and entities can also contribute to the generation of meaningful and coherent responses, especially in knowledge-aware CRS, see \cite{chen2021knowledge,wang2013perspectives,zhou2020improving,10.1145/3475959.3485392}.

Annotating items and entities can be a laborious and economically expensive process \cite{arjannikov2013verifying,grosman2020eras}. Human costs are high and may even be prohibitive for domains where particular knowledge or expertise are required to accomplish the annotation task \cite{benato2021semi}. In that context, the quality of the resulting annotations is crucial, and factually wrong annotations can lead to errors or ambiguity for the downstream task. Automating the annotation task or at least automatically verifying the annotations \cite{arjannikov2013verifying} has therefore been in the focus of research for several years.

We note here that data quality is crucial both for recent \emph{generation-based} CRS approaches as well as for \emph{retrieval-based} approaches to build natural language conversational systems \cite{manzoorrs2021}. For both types of systems, the question arises to what extent better data quality, i.e., having correct annotations and noise-free conversations, leads to better results in terms of the quality of the responses returned by a system for a given user utterance, e.g., in terms of consistency and plausibility.


In this work, we study the recent \inspmain dataset, in which the items and entities that were mentioned in the recorded utterances are explicitly annotated. 
These annotations were created with the help of automatic approaches using keyword or pattern matching methods. However, looking at the data, we observed a substantial number of cases where items and entities were either wrongly annotated or missing annotations at all, e.g., \emph{``My favorite [ MOVIE\_GENRE\_1] \xspace are Groundhogs Day, [ MOVIE\_TITLE\_2] \xspace and Borat''}. In addition, there were several cases where the utterances included noise, e.g., \emph{``How did you like QUOTATION\_MARKHustlersQUOTATION \_MARK?''}. Finally, we found instances where regular words were identified as being named entities. In this latter case, human annotations would in fact have been required.\footnote{Consider the movie ``It (2017)'' as an example of a difficult case, e.g., when appearing in an utterance like ``Have you seen It?''.} Overall, such issues may limit the quality of any CRS that is built on top of such data.

To understand the severity of the problem and the potential effects of data issues on the quality of a CRS, we have \emph{manually} corrected the dataset by fixing the annotations and by removing noise from the utterances. Then, we conducted offline experiments and human evaluations to compare the performance of different benchmark CRS when using the original (\inspmain) and improved (\inspcurrent) datasets. Overall, the results of our analyses indicate that all CRS showed better performance in different dimensions when built on \inspcurrent. In order to facilitate the design and development of future sociable CRS, we release the \inspcurrent dataset online at \url{https://github.com/ahtsham58/INSPIRED2}.



\section{Related Work}
\label{sec:work}

In this section, we first discuss datasets and aspects of data quality in the context of CRS. Afterwards, we review different design paradigms for building CRS, followed by a discussion of predominant evaluation approaches for such systems.

\paragraph{Datasets and Data Quality}
Research interest in CRS has experienced a substantial growth in recent years, see \cite{jannach2021crscsur,gao2021advances} for related surveys. Many current systems interact with users in natural language, and one important goal for such system is to enable them to engage in conversations that reflect human behavior. Since many of these recent systems are built on recorded dialogs between humans, the capabilities of the resulting CRS depend on the richness of the communication in the datasets, e.g., in terms of the user \emph{intents} that can be found in the conversations, see~\cite{CaiChen2020UmapIntents} for an detailed analysis of such intents.

A number of new datasets for conversational recommendation were published in recent years, e.g., \cite{li2018towards,kang2019recommendation,zhou2020towards,fu2020cookie}. Such datasets, which are commonly collected with the help of crowdworkers, can however have limitations and may be not fully representative in terms of what we would observe in reality. In some cases, for example, crowdworkers were instructed to mention a minimum number of movies in the conversations. This leads to mostly ``instance-based'' conversations, where crowdworkers rather mention individual movies they like than their preferred genres, see also, \cite{li2018towards,christakopoulou2016towards,ren2021learning}.

Another problem when creating such datasets lies in the recognition and annotation of named entities appearing in the conversations, as mentioned above. Annotating entities in textual data can be a tedious process that may require a substantial amount of manual effort and time. To overcome this challenge, researchers sometimes adapt a semi-automatic approach or rely on NLP-assisted tools that visualize the entities in a text in order to reduce the required manual effort \cite{benato2021semi,grosman2020eras,stenetorp2012brat}. Generally, some automatic approaches may experience problems to correctly create annotations because human judgments and opinions are required. An automated approach was used in the context of the \inspmain dataset. Here, the items and entities were annotated using keyword or pattern matching approaches. However, verifying the outcomes of such automatic or semi-automatic approaches can again be laborious and require manual effort. 

Today, structured annotations for items and entities mentioned in the conversations are common in recent datasets. For example, in the case of the ReDial dataset \cite{li2018towards}, the mentioned movie titles were annotated with unique IDs. However, the ReDial dataset has some limitations. Various meta-data concepts (e.g., genres, actors, or directors) were not annotated. Moreover, the recorded dialogs include limited social interactions or explanations for the made recommendations. On the other hand, the \inspmain dataset includes rich sociable conversation and explanation strategies for the recommended items. Also, aspects like movie genres or actors were explicitly annotated too. A comparison of these differences can be found in \cite{hayati2020inspired}. 
The key statistics of the \inspmain dataset are shown in Table~\ref{tab:INSPIRED2-statistics}.

As mentioned earlier, the \inspmain dataset has some limitations. The keyword or pattern matching approach used for the annotations might for example not detect misspelled keywords or concepts in an utterance. Moreover, data anomalies such as noisy utterances or ill-formed language can deteriorate the performance of an annotating algorithm, leading to challenges for the downstream use of the dataset \cite{grosman2020eras,Zong2021,rottger2021two}. In reality, the level of noise can be substantial both in real-world applications and in purposefully created datasets. Therefore, data quality assurance is often considered a significant and important step in NLP applications.

\begin{table}[h!t]
  \caption{Main Statistics of \inspmain}
  \centering
 \begin{tabular}{p{5.5cm}c}
  \toprule
  & \textbf{Total}\\ \toprule
  Number of dialogs (conversations)   & 1,001 \\
  Average turns per dialog &  10.73 \\
  Average tokens per utterance &  7.93 \\
  \bottomrule
  Number of human-recommender utterances & 18,339 \\
  \bottomrule
  Number of seeker utterances & 17,472 \\
  \bottomrule
  \end{tabular}
  \label{tab:INSPIRED2-statistics}
\end{table}



\paragraph{Building Conversational Recommender Systems}
Research on CRS has made substantial progress in terms of their underlying technical approaches. Some early commercial system such as \emph{Advisor Suite}~\cite{Jannach:2004:ASK:3000001.3000153} for example relied on an entirely knowledge-based approach for the development of adaptive and personalized applications. Similarly, early critiquing-based systems were based on detailed knowledge about item features and possible critiques and had limited learning capabilities~\cite{mccarthy2010experience,chenpucritiquing2012}.


Technological advancements particularly in fields like NLP, speech recognition, and machine learning in general led to the design of today's end-to-end learning-based CRS. In such approaches, recorded recommendation dialogs between paired humans are used to train the deep neural models, see, e.g., \cite{chen-etal-2019-towards,zhou2021crfr,chen2021knowledge,zhou2020improving}. Given the last \emph{user} utterance and the history of the ongoigng dialog history, these trained models are then used to \emph{generate} responses in natural language. These responses can either include item recommendations, which are also computed with the help of machine learning techniques, or other types of conversational elements, e.g., greetings.

In terms of the underlying data, the DeepCRS \cite{li2018towards} system was built on the ReDial dataset, which was created in the context of this work. Later on, systems were developed which also relied on this dataset as well but included additional information sources, e.g., from \emph{DBPedia} or \emph{ConceptNet} \cite{chen2019towards,zhou2020improving}, to build knowledge graphs that are then used to improve the generated utterances. A number of works also makes use of pretrained language models like BERT \cite{Devlin2019BERTPO} and subsequently fine-tune them using the recommendation dialogs, see, e.g., \cite{wang2021finetuning}. A related approach was adapted by the authors of \inspmain, in which they proposed two variants of a conversational system, \emph{with} and \emph{without} strategy labels. 


Unlike generation-based systems, in retrieval-based CRS the idea is to retrieve and adapt suitable responses from the dataset of recorded dialogs. One main advantage of retrieval-based approaches is that the retrieved responses were genuinely made by humans and thus are grammatically usually correct and in themselves semantically meaningful \cite{MANZOOR2021100139}. Recent examples of such retrieval-based systems are \rbCRS \cite{manzoorrs2021} and \crbCRS \cite{manzoor2021towardscrbcrs}, which we designed and evaluated based on the ReDial dataset in our own previous work. 

\paragraph{CRS Evaluation}
Evaluating a CRS is a multi-faceted and challenging problem as it requires the consideration of various quality dimensions. An in-depth discussion of evaluation approaches for CRS can be found in \cite{jannach2022evaluatingcrs}. Like in the recommender systems literature in general, computational experiments that do not involve humans in the loop are the predominant instrument to assess the quality of a CRS. Common metrics to evaluate the quality of the recommendations include \emph{Recall}, \emph{Hit Rate}, or \emph{Precision} \cite{chen-etal-2019-towards,kang2019recommendation,zhang2021kecrs}. Moreover, certain linguistic aspects such as fluency or diversity are often evaluated with offline experiments as well to assess the quality of the generated responses. Common metrics in this area include \emph{Perplexity}, \emph{distinct N-Gram}, or the \emph{BLEU} score \cite{li2018towards,chen-etal-2019-towards,10.1145/3475959.3485392,zhou2020towards}.




Given the interactive nature of CRS, offline experiments and the corresponding metrics have their limitations. Mainly, it is not always clear if the results obtained from offline experiments are representative of the user-perceived quality of the recommendations or system responses in general \cite{MANZOOR2021100139}. For example, when using metrics like the \emph{BLEU} score, usually a system response is compared with one particular given ground truth. Such a comparison has limitations when used to estimate the average quality of a system's responses, because there might be many different alternative responses that might be suitable as well in an ongoing dialog. Still, offline evaluations have their place and value. They can for example be informative for assessing particular aspects such as the number of items or entities that appear in an utterance or conversation. 


Overall, given the limitations of pure offline experiments, researchers often follow a mixed approach where some aspects of the system are evaluated offline and some with humans. Typical quality aspects in terms of human perceptions in such combined approaches include the assessment of the \emph{meaningfulness} or \emph{consistency} of the system responses \cite{hayati2020inspired,chen-etal-2019-towards,zhou2020improving,10.1145/3475959.3485392,manzoor2021towardscrbcrs}.



\section{Data Annotation Methodology}\label{sec:di-methodology}
During the creation of the \inspmain \cite{hayati2020inspired} dataset, items and other entities were annotated in an automated way, as described above.
For example, genre keywords were annotated using a regular expression to match a set of predefined tokens. Regarding actors and directors and other entities, a pattern matching technique was used, where words starting with a capital letter were searched in the TMDB database\footnote{\url{https://www.themoviedb.org}}. A similar technique was used for movie titles. However, as mentioned, we observe a large number of cases where items and entities were either wrongly annotated or missing annotations.
To answer our research question on the impact of the quality of the underlying data on the quality of the responses of a CRS, we fixed the annotations as follows.

\paragraph{Procedure}
To fix the annotations, we interviewed a number of university students to assess their knowledge in the movies domain and their ability to do the correction task. Subsequently, we hired two students and instructed them on how to annotate and clean the dataset. First, they were briefed on the logical format of the original annotations and how to retain that format. Second, they were asked to read each utterance individually, to detect potential noise, and to analyze which items or entities (e.g., title, genre, actor, or director) are mentioned in it.

In case of ambiguity or obscurity, they were allowed to access online portals, e.g., IMDb\footnote{\url{https://www.imdb.com/}}. Note that regarding the genres, a set of 27 keywords was provided to them, which we curated and used in our earlier research \cite{manzoor2021towardscrbcrs}. 
After the briefing, the dataset was split evenly for both annotators. On weekly basis, their performance and the accuracy of the annotations was checked by one of the authors. Finally, after annotating the complete dataset, a number of additional validation steps were applied.


First, using a Python script, we ensured that every placeholder is enclosed by `[' and `]' as was done originally, e.g.,  [MOVIE\_TITLE\_1]. Second, another thorough manual examination of the entire improved dataset was performed to fix any missing annotations or noise. In that context, we also double-checked the consistency of the format and of the annotations.

\paragraph{The \inspcurrent Dataset}
In total, 1,851 new annotations were added to \inspmain, leading to the \inspcurrent dataset. The most mistakes or inconsistencies were found for the items, i.e., movie titles, which is the most pertinent information for developing a CRS. We present the statistics about \emph{new} annotations in Table \ref{tab:new-annotation}. Overall, we added around 20\% new annotations in \inspcurrent. 
The number of issues that were fixed, e.g., duplicate annotations in an utterance, noise or factually wrong information in the original annotations, are not shown in the presented statistics. We release the \inspcurrent both in the TSV and JSON format {\href{https://github.com/ahtsham58/INSPIRED2/tree/main/Dataset}{online}.

\begin{table}[h!t]
  \caption{Statistics about new annotations added in \inspcurrent}
  \centering
 \begin{tabular}{p{4cm}cc}
  \toprule
  & \textbf{Total} & \textbf{\% Increase}\\ \toprule
  Number of movie titles  & 966 & 22.0\\
  Number of movie genres &  206 & 5.0 \\
  Number of actors, directors, etc. &  519 & 49.0 \\
  Number of movie plots &  160 & 54.6\\
  \bottomrule
  Number of new annotations &  1851 & 18.9 \\
  \bottomrule
  \end{tabular}
  \label{tab:new-annotation}
\end{table}

\paragraph{Observed Issues}
During the annotation process, we recorded the observed issues in the original annotations. Since the original annotations were created using automatic techniques, many issues were related to the limitations of the simple keyword or pattern matching techniques. Overall, we observed a number of cases where minor spelling mistakes or incomplete movie titles made the exact string matching approaches ineffective. 

For example, in one of the utterances, ``\emph{I think I am waiting for Star Wars The Rise of Skywalker''}, the annotation was missing because the correct title is ``\emph{Star Wars: Episode IX – The Rise of Skywalker}''. Similarly, we observe a significant number of cases where an utterance was only partially annotated, e.g., ``\emph{ok is it scary like incidious or [ MOVIE\_GENRE\_2] [ MOVIE\_TITLE\_5]}''. In addition, at places where two entities were separated with `/' instead of a space, the automatic technique often failed to create proper annotations, e.g., ``\emph{Since you like [MOVIE\_GENRE\_1] drama/mystery, I'm going to send you the trailer to the movie [MOVIE\_TITLE\_3]}''.

Also, the automatic approach used for \inspmain sometimes had difficulties to deal with ambiguity. We found a number of cases where a regular word was annotated, although such a word did not belong to any item or entity. For example, in one of the cases, ``\emph{Are you interested in a current movie in the box office?}'', the utterance was annotated as ``\emph{Are you interested in a current movie in the box [ MOVIE\_TITLE\_0]}'', where the word `\emph{office}' was mistakenly annotated as an item, i.e., \emph{The Office (2005)}.

Overall, the main observed issues are the following.
\begin{enumerate}
    \item Missing annotations for movie titles, genres, actors, movie plots, etc.
    \item Partially annotated items and entities such as movie titles, or genres in an utterance.
    \item Factually wrong annotations for movie titles.
    \item Inconsistent indexing for the annotated items and entities.
    \item Mistaken annotations for plain text, e.g., \emph{family, box office}; human annotations may be required here.
    \item Parts of the utterance or a few keywords were omitted during the annotation process.
\end{enumerate}

\section{Evaluation Methodology}
\label{sec:exp-design}
We performed both offline experiments as well as a human evaluation to assess the impact of data quality on the quality of the responses of a CRS.

\paragraph{Offline Evaluation of Recommendation Quality}
We included the following recent end-to-end learning approaches in our experiments: DeepCRS \cite{li2018towards}, KGSF \cite{zhou2020improving}, TG-ReDial \cite{zhou2020towards}, and the INSPIRED model \emph{without strategy labels}\footnote{The INSPIRED \emph{with strategy labels} model was not publicly available.} \cite{hayati2020inspired}. This selection of models covers various design approaches for CRS, e.g., using an additional knowledge graph or not. We used the open-source toolkit CRSLab\footnote{\url{ https://github.com/RUCAIBox/CRSLab}} for our evaluations. This framework was used in earlier research as well, for example in \cite{chen2021knowledge,zhou2022c2,li2021knowledge}. For our analyses, we first trained the aforementioned CRS models using the original split ratio, i.e., 8:1:1, for each dataset. 
Afterwards, given the trained models and test data for each dataset, we ran three trials for each CRS and subsequently averaged the results for offline evaluation metrics. Note that the same procedure was adapted for both versions of the dataset, i.e., \inspmain and \inspcurrent.


\paragraph{User Study on Linguistic Quality}
We conduct a user study to compare the perceived quality of system responses using either \inspmain and \inspcurrent. Specifically, we randomly sampled \emph{same} 50 dialog situations from each dataset. To create the dialog continuations, we used the \emph{retrieval-based} CRS approaches, \rbCRS and \crbCRS, which we proposed in our earlier work, see \cite{manzoor2021towardscrbcrs}.

In order to obtain fine-grained assessments, three human judges\footnote{These judges were PhD students and were different than the ones who fixed the annotations.} were involved. The specific task of the judges was to assess (rate) the \emph{meaningfulness} of a system response as a proxy of its \emph{quality} and \emph{consistency} in a dialog situation, see \cite{li2018towards,JannachManzoor2020,zhou2020improving}. Note that in this study we did not explicitly assess the quality of the specific item recommendations. Instead, the focus of this study was to understand the impact of the improved underlying dataset on the linguistic quality and the consistency of the generated responses. 

We used a 3-point scale for these ratings, from `\emph{Completely meaningless (1)}' to `\emph{Somewhat meaningless and meaningful (2)}' to `\emph{Completely meaningful (3)}'. The human judges were provided with specific instructions on how to evaluate the meaningfulness of a response, e.g., they should assess if a response represents a logical dialog continuation and evaluate the overall language quality of the given response.
Overall, the human judges were provided 50 dialogs (446 responses to rate) that were produced using the \inspmain and \inspcurrent datasets. 
We also explained the meanings and purpose of various placeholders contained in the responses to the human judges. Moreover, to avoid any bias in the evaluation process, the judges were not made aware which response was created for which dataset by which CRS.
Also, the order of the dialogs and the system responses were randomized. 

\section{Results}
\label{sec:results}
\paragraph{Recommendation Quality}
Table \ref{tab:results-rec} shows the accuracy results for the evaluated CRS models. Specifically, we provide the results for the different benchmark CRS models in terms of the performance \emph{difference} when using the original and improved annotations. Overall, we can observe an almost consistent gain in performance for all models and on all metrics except Hit@50 when the improved dataset is used. The obtained improvements can be quite substantial, indicating that improved data quality can be helpful for CRS of different types, including \emph{(i)} CRS, which do not rely on additional knowledge sources, \emph{(ii)} CRS that leverage additional knowledge sources, \emph{(iii)} CRS that are guided by a topic policy, and \emph{(iv)} CRS that rely on pre-trained language models like BERT.

Interestingly, we see negative effects for two measurements in which Hit@50 is used as a metric. A deeper investigation of this phenomenon is needed, in particular as the other metrics at this (admittedly rather uncommon) list length, MRR@50 and NDCG@50, indicate that the improved dataset is helpful to increase recommendation accuracy. At the moment, we can only speculate that the improved annotations in the ongoing dialog histories led to more diverse or niche recommendations compared to the original dataset. We might assume that the missing annotations in many cases referred to less popular movies, so that the recommendations without the improved annotations will more often recommend popular movies, which is commonly advantageous in terms of hit rate and recall.



\paragraph{Linguistic Quality}

We recall that three human evaluators assessed the linguistic quality of the system responses (dialog continuations), which were created either based on the \inspmain or the \inspcurrent dataset. As underlying CRS systems, we considered the retrieval-based approaches \rbCRS and \crbCRS, as mentioned above. For our analysis, we averaged the scores by the three evaluators. Table \ref{tab:results-user-study} shows the mean ratings across all dialog situations as well as the standard deviations. We find that also in the case of retrieval-based approaches, improving the quality of the underlying dataset was helpful, leading to higher mean scores, without observing larger standard deviations. A Student's t-test reveals that the observed differences in the means are statistically significant (\emph{p<0.001}). \footnote{We provide the data and compiled results of our study  \href{https://github.com/ahtsham58/INSPIRED2/tree/main/Results}{online}.}

\begin{table*}[ht!]
\caption{Accuracy results obtained in the \emph{offline} evaluation. V1 represents \inspmain,  V2 denotes \inspcurrent, and ``\% Change'' represents the actual performance gain/loss when using \inspcurrent compared to \inspmain.}
\centering
\begin{adjustbox}{max width=\textwidth}
\small
\begin{tabular}{lcccccccccc}
\toprule
    &  & \textbf{Hit@1} & \textbf{Hit@10} & \textbf{Hit@50} & \textbf{MRR@1} & \textbf{MRR@10} & \textbf{MRR@50} &\textbf{NDCG@1} & \textbf{NDCG@10}&\textbf{NDCG@50}\\  \hline

 \multirow{3}{1.5cm}{DeepCRS \cite{li2018towards}} & V1 & 0.0006  & 0.0464 &0.1726 &0.0065 & 0.0148& 0.0193 &0.0065 &0.0220	  & 0.0478 \\
 & V2 & 0.0256  & 0.0578 &0.1222  &0.0256 & 0.0306 & 0.0333 &0.0256 &0.0366  &0.0504 \\
 \cline{2-11}
  & \textbf{ \% Change} & \textbf{4161.11}  & \textbf{24.50} & -29.20  & \textbf{294.95} & \textbf{106.99} & \textbf{72.48} & \textbf{294.95} & \textbf{66.08}  &\textbf{5.31} \\
\bottomrule

 \multirow{3}{1.5cm}{KGSF \cite{zhou2020improving}} & V1 & 0.0022  & 0.0216 & 0.0744 & 0.0032 & 0.0061 & 0.0084 & 0.0022 & 0.0097 & 0.0211  \\
& V2 & 0.0066  & 0.0303 & 0.0587 & 0.0057 & 0.0123 & 0.0134 & 0.0066 & 0.0165 & 0.0223\\
 \cline{2-11}
  & \textbf{ \% Change} & \textbf{207.27}  & \textbf{40.46} & -21.11  & \textbf{75.58} & \textbf{100.97} & \textbf{58.40} & \textbf{207.27} & \textbf{70.36}  &\textbf{5.78} \\
\bottomrule

 \multirow{3}{1.5cm}{TG-ReDial \cite{zhou2020towards}} & V1 & 0.0365  & 0.1149 & 0.2344 & 0.0365 & 0.0572 & 0.0626 & 0.0365 & 0.0707 & 0.0967 \\
& V2 & 0.0511  & 0.1315 & 0.2417 &0.0511 & 0.0742 &0.0792 &0.0511 &0.0877  &0.1118\\
 \cline{2-11}
    & \textbf{ \% Change} & \textbf{40.00}  & \textbf{14.46} & \textbf{03.12}  & \textbf{40.00} & \textbf{29.64} & \textbf{26.48} & \textbf{40.00} & \textbf{24.05}  &\textbf{15.51} \\
 \bottomrule

 \multirow{3}{1.5cm}{INSPIRED \cite{hayati2020inspired}} & V1 & 0.0151  & 0.0550 & 0.1532 & 0.0151& 0.0241 &0.0286 & 0.0151 & 0.0312 & 0.0527 \\
& V2 & 0.0194  & 0.0734 & 0.1855 & 0.0194& 0.0293 & 0.0353&0.0194 &0.0392  & 0.0650\\
 \cline{2-11}
    & \textbf{ \% Change} & \textbf{28.57}  & \textbf{33.33} & \textbf{21.13}  & \textbf{28.57} & \textbf{21.59} & \textbf{23.44} & \textbf{28.57} & \textbf{25.44}  &\textbf{23.28} \\

\bottomrule
\end{tabular}
\end{adjustbox}
\normalsize
\label{tab:results-rec}
\end{table*}



\begin{table}[h!t]
\caption{Results of Human Evaluation}
\centering
\begin{tabular}{lccc}
\toprule
    &  & \textbf{\inspmain} & \textbf{\inspcurrent} \\  \hline
 \multirow{2}{1.0cm}{\rbCRS} & Average score & 2.30  & \textbf{2.46} \\
 & Std. deviation & 0.62  & \textbf{0.59} \\
 \hline
\multirow{2}{1.0cm}{\crbCRS} & Average score & 2.31  & \textbf{2.46} \\
& Std. deviation & 0.55  & \textbf{0.55} \\
\bottomrule

\end{tabular}
\label{tab:results-user-study}
\end{table}

\paragraph{Comparison of Knowledge Concepts in Responses}
To understand the impact of the new annotations on the responses in terms of the richness of knowledge concepts, we compute the number of items and entities that appeared in the system responses. Specifically, we compute the number of placeholders in the responses, before they would be replaced by the recommendation component, see \cite{manzoor2021towardscrbcrs}. In Table \ref{tab:entities}, we present the statistics for \rbCRS and \crbCRS for both dataset versions. Overall, we find that the responses for the improved dataset contain between 20\% and 27\% more concepts and entities. We note that an increase in concepts is expected, as \inspcurrent has almost 20\% more annotations. However, the important observation here is that the retrieval-based CRS approaches actually surfaced these richer system responses frequently.

\begin{table}[h!t]
\caption{Number of Items and Entities included in Responses}
\centering
\begin{tabular}{lccc}
\toprule
    &   \textbf{\inspmain} & \textbf{\inspcurrent} & \textbf{\% Increase} \\  \hline
 \rbCRS & 174  & \textbf{222} & \textbf{27.6}\\
\crbCRS &  208  & \textbf{251} & \textbf{20.7} \\
\bottomrule

\end{tabular}
\label{tab:entities}
\end{table}

\paragraph{BLEU Score Analysis}
Finally, in order to understand to what extent (offline) linguistic scores correlate with the perceived quality of responses as was done in \cite{hayati2020inspired,chen-etal-2019-towards}, we performed an analysis of the BLEU scores obtained for the different datasets. Specifically, given a system response and the corresponding \emph{ground truth} response, we preprocess both sentences and compute the BLEU scores for $N=\{1,2,3,4\}$ grams. We provide the results of this analysis online. In sum, the analysis shows that the BLEU scores generally improve when the underlying data quality is higher, i.e., in the case of the \inspcurrent dataset. These findings are thus well aligned with the outcomes of our human evaluation study, where using \inspcurrent as an underlying dataset turned out to be favorable.

\section{Conclusion}
\label{sec:conclusion}
Datasets containing recorded dialogs between humans are the basis for many modern CRS. In this work, we have analyzed the recent \inspmain dataset, which was developed to build the next generation of sociable CRS. We found that automatic entity and concept labeling has its limitations and we have improved the quality of the dataset through a manual process. We then conducted both computational experiments as well as experiments with users to analyze to what extent improved data quality impacts recommendation accuracy and the quality perception of the system's responses by users. The analyses clearly indicate the benefits of improved data quality across different technical approaches for building CRS.
We release the improved dataset publicly and hope to thereby stimulate more research in sociable conversational recommender systems in the future.

\newpage
\bibliography{main}
\balance
\end{document}